\documentclass[runningheads]{llncs}

\usepackage{eccv}

\usepackage{eccvabbrv}

\usepackage{graphicx}
\usepackage{booktabs}
\usepackage[subpreambles=true]{standalone}

\usepackage[accsupp]{axessibility}  %

\usepackage{hyperref}

\usepackage{orcidlink}

\newcommand{\PAR}[1]{\paragraph{#1}}

\newcommand{\arx}[1]{\textcolor{black}{#1}}

\begin{document}

\title{IntrinsicAnything: Learning Diffusion Priors \\ for Inverse Rendering Under \\ Unknown Illumination} 

\titlerunning{IntrinsicAnything}

\author{Xi Chen\inst{1} \and
Sida Peng\inst{1} \and
Dongchen Yang\inst{1} \and
Yuan Liu\inst{2} \and \\
Bowen Pan\inst{3} \and
Chengfei Lv\inst{3} \and
Xiaowei Zhou\inst{1}
}
\authorrunning{Chen. et al.}

\institute{State Key Lab of CAD\&CG, Zhejiang University \and The University of Hong Kong \\ \and
Tao Technology Department, Alibaba Group
}
\maketitle

\begin{abstract}
This paper aims to recover object materials from posed images captured under an unknown static lighting condition.
Recent methods solve this task by optimizing material parameters through differentiable physically based rendering.
However, due to the coupling between object geometry, materials, and environment lighting, there is inherent ambiguity during the inverse rendering process, preventing previous methods from obtaining accurate results.
To overcome this ill-posed problem, our key idea is to learn the material prior with a generative model for regularizing the optimization process.
We observe that the general rendering equation can be split into diffuse and specular shading terms, and thus formulate the material prior as diffusion models of albedo and specular.
Thanks to this design, our model can be trained using the existing abundant 3D object data, and naturally acts as a versatile tool to resolve the ambiguity when recovering  material representations from RGB images.
In addition, we develop a coarse-to-fine training strategy that leverages estimated materials to guide diffusion models to satisfy multi-view consistent constraints, leading to more stable and accurate results.
Extensive experiments on real-world and synthetic datasets demonstrate that our approach achieves state-of-the-art performance on material recovery. The code will be available at \href{https://zju3dv.github.io/IntrinsicAnything/}{https://zju3dv.github.io/IntrinsicAnything/}.
\end{abstract}
    
\section{Introduction}
\label{sec:intro}

Recovering the object's geometry, material, and lighting from captured images, also known as inverse rendering, is a long-standing task in computer vision and graphics. 
These physical attributes of 3D objects are essential for many applications, such as VR/AR, movie production, and video games.
Due to the inherent complexity of interaction between real-world objects and environment lighting, inverse rendering remains an ill-posed problem.
Previous works overcome this problem using sophisticated capture systems~\cite{10.1145/344779.344855, 10.1145/3355089.3356571}, or co-located flashlight and camera in a dark environment~\cite{MobileSVBRDF:SIGA:2018, bi2021avatar, zhang2022iron}. However, these method requires special hardware equipment or constrained environment, limiting their applications.

Based on differentiable physically based rendering and neural scene representations, recent methods~\cite{nerfactor, physg2021, zhang2022invrender, hasselgren2022nvdiffrecmc, Jin2023TensoIR, zhang2023neilfpp} are able to object materials from images captured under natural lighting.
They exploit neural networks to represent object materials and geometry, and combine these object attributes with the learnable lighting to synthesize images, which are compared with captured images to optimize the model paramaters.
To achieve better performance of inverse rendering, these methods have attempted to improve object representations~\cite{physg2021,Jin2023TensoIR}, lighting representations~\cite{hasselgren2022nvdiffrecmc,zhang2023neilfpp}, and training strategies~\cite{zhang2022invrender,nerfactor}.
However, we argue that depending solely on the rendering loss is insufficient for accomplishing accurate decomposition, due to the inherent ambiguity in the inverse rendering process.

\begin{figure}[t!]
    \vspace{0cm}
    \centering
       \includegraphics[width=1\linewidth]{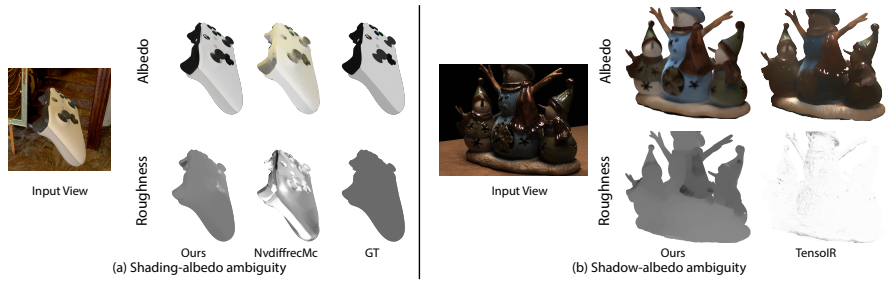}
       \caption{
           \textbf{Two types of ambiguities in inverse rendering.}
            (a) Ambiguity between diffuse shading and albedo.
            For example, the Xbox is lit by a yellow light, and the decomposed albedo from NVdiffrecMC~\cite{2022nvdiffrecmc} tends to be yellow.
            (b) Ambiguity between shadow and albedo.
            For example, the porcelain toy is with self-occlusion, and TensoIR~\cite{Jin2023TensoIR} bakes the shadow into the recovered albedo.
            Our method well handles the two types of ambiguities.
       }
       \label{fig:intro-amb}
\end{figure}

We observed that there are two categories of ambiguities, as shown in Fig.~\ref{fig:intro-amb}.
First, there is ambiguity between diffuse shading and albedo.
Intuitively, when a white object is lit by a yellow light, the decomposed albedo appears to be yellow.
Second, when an object is with self-occlusion and illuminated under a complex lighting condition, the hard-cast shadow can be easily baked into the albedo.
Due to these ambiguities, the optimization process of previous methods tend to gets trapped in local optimal, leading to the artifacts on the recovered material.

\begin{figure}[t!]
    \vspace{-1cm}
    \centering
       \includegraphics[width=1.0\linewidth]{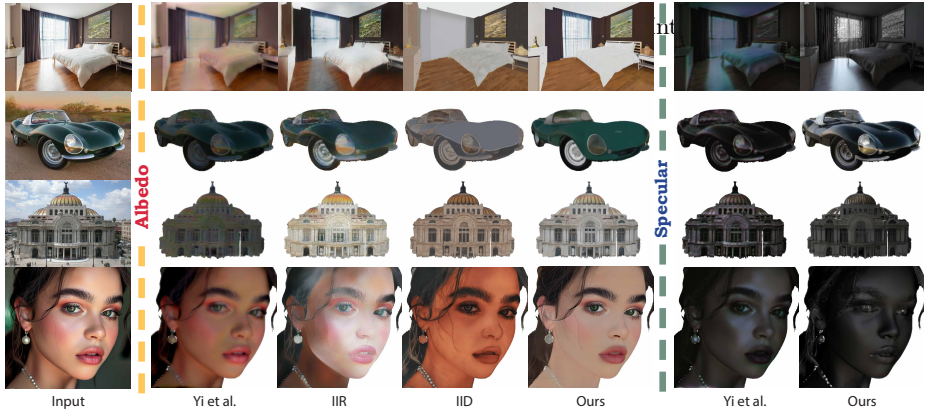}
       \caption{
           \textbf{Single-view intrinsic images decomposition results}. \arx{Compared with the objects-level method Yi \etal \cite{yi2023weaklysupervised} as well as scene-level methods IIR~\cite{zhu2022learning} and IID~\cite{kocsis2023iid}, our approach recovers more accurate and detailed intrinsic images and demonstrates strong generalization capabilities across various objects and scenes.}
       }
       \label{fig:teaser}
\end{figure}

In this paper, we seek to use data-driven priors for ambiguity-free inverse rendering. Our key idea is learning the distribution of albedo and specular shading with generative models, which is then used to regularize the optimization process.
Inspired by the recent progress of generative models\cite{rombach2021highresolution}, we exploit the conditional diffusion model for modeling the material prior.
Our choice of material prior for albedo and specular shading is motivated by the fact that the general rendering equation can be separated into diffuse and specular terms, according to the Disney BRDF model, which thus has two advantages.
First, we can construct training data on albedo and specular using a large amount of 3D objects~\cite{objaverse} that contain various BRDF models.
\arx{
This enables our prior model to generalize across different domains, as shown in Fig.~\ref{fig:teaser}.
}
Second, our designed prior can work as a versatile tool to resolve the ambiguity in the inverse rendering process of many material representations in theory.

A challenge of applying diffusion models to regularize the optimization process is that such generative model struggles to perform multi-view consistent constraints.
To address this problem, we propose a coarse-to-fine optimization scheme.
Our approach first leverages the diffusion prior to recover the coarse object material and environment lighting.
Then, the coarse albedo and specular shading are used to guide the diffusion model to produce constraints with better multi-view consistency, leading to more accurate material and lighting recovery.

For training our model, we create a dataset consisting of RGB, albedo, and specular images based on the Objaverse dataset~\cite{objaverse}.
To validate the effective of the proposed approach, we conduct extensive experiments on synthetic and real-world data. Our method achieves state-of-the-art performance on several benchmarks, as well as generalizing to internet images, as shown in Fig.~\ref{fig:teaser}.

\section{Related Work}
\label{sec:related}

\PAR{Learning priors for inverse rendering.}
Earlier approaches for inverse rendering usually require controlled set-ups~\cite{Levoy2000TheDM,debevec2000acquiring} and strong domain specific assumptions to reconstruct scene properties from images~\cite{10.1145/280814.280874}.
With the advance of deep learning, the intrinsic properties of images can be directly learned from large-scale synthetic datasets~\cite{Shi2017CVPR,li2021openrooms,Li2018sigasia, zhu2022learning}.
Shi \etal~\cite{Shi2017CVPR} leverages ShapeNet~\cite{shapenet} and introduces a dataset with image intrinsics to decompose the image into base color and shading.
Li \etal~\cite{Li2018ECCV} introduces a large-scale scene dataset and lifts the model's capability to from object level to scene level. Subsequent works~\cite{lichy2021,li2018learning,sang2020single} have attempted to further recover the BRDF and geometry of a scene. Recently, Kocsis \etal~\cite{kocsis2023iid} utilizes diffusion models to estimate the intrinsics of single view indoor scene images and demonstrates a significant improvement on the generalization from synthetic images to real images. 
However, these learning based methods are largely constrained by the photorealism of synthetic images because the ground truth material properties and illumination for real scenes is prohibitively difficult to obtain.
To mitigate the lack of labeled data issue,~\cite{yu19inverserendernet,YuSelfRelight20,YuSelfRelight20,wimbauer2022derendering,Liu2020unsupervised} uses unsupervised methods and introduces various physically and statistically motivated constraints to achieve training without ground truth.~\cite{yu19inverserendernet} takes multiple images as input and introduces multi-view stereo supervision to achieve training a single view inverse rendering model without ground truth supervision.~\cite{YuSelfRelight20} decomposes outdoor scenes via a physically based rendering constraint.~\cite{Liu2020unsupervised} models inverse rendering as an image translation problem and learns to map the image from the image domain to the intrinsics domain. Although these supervised and unsupervised learning based methods achieves promising progress on inverse rendering, their inference on images from the same scene usually lacks of consistency. In contrast, our method combines the monocular diffusion models with the multiview physically based rendering and thus achieved accurate material and illumination estimation with multiview consistency.

\PAR{Inverse rendering with neural representation.}

Inspired by the implicit representations of NeRF~\cite{mildenhall2020nerf}, various works~\cite{boss2021nerd, physg2021, Nefactor2021, Jin2023TensoIR, liu2023nero} extend the neural representations to factorize a scene into material, geometry, and lighting.
NeRV~\cite{nerv2020} leverages a neural visibility field to reduce the computational cost and models indirect lighting with known environment lighting conditions.
InvRender~\cite{2022invrender} and NeILF series~\cite{yao2022neilf, zhang2023neilf++} further model indirect illumination with a neural representation at unknown light conditions.
NvdiffrecMC~\cite{2022nvdiffrecmc} and NeFII~\cite{wu2023nefii} incorporate Monte Carlo ray tracing to recover high-frequency details of lighting. 
~\cite{sun2023neuralpbir} adopts a coarse-to-fine manner to retrieve the materials and illumination of a scene.
\arx{
More recently, GS-IR~\cite{liang2023gs} and GIR~\cite{shi2023gir} leverage Gaussian Splatting~\cite{kerbl3Dgaussians} to efficiently modeling indirect lighting for inverse Rendering.
}
Furthermore,~\cite{zhu2023i2, fipt2023} attempt to scale up the neural representations for solving inverse rendering from object to indoor scene. However, due to the inherit ill-posed feature of inverse rendering problem, these methods are still suffered from disentangling the materials and illumination properties from images. In contrast, by learning a strong material and shading prior, we effectively tackled the ill-posed feature and received a significant improvement on decomposing the image into BRDFs and illuminations. 

\PAR{Diffusion models.}
Recently, diffusion models have demonstrated unprecedented performance on generation tasks, such as 2D generation~\cite{croitoru2023diffusion, ho2020denoising, rombach2022high, saharia2022photorealistic} and 3D generation~\cite{poole2022dreamfusion, tang2023mvdiffusion, anciukevivcius2023renderdiffusion, chen2023single, cheng2023sdfusion, liu2023syncdreamer}, indicating the impressive capability of learning the distribution of target data.
Inspired this, some recent methods~\cite{kawar2022denoising, chung2022come} have attempted to exploit the diffusion prior for ill-posed inverse problems, including image denoising~\cite{xie2023diffusion, zhu2023denoising}, image completion~\cite{horita2022structure, saharia2022palette}, and single-view reconstruction~\cite{Tang_2023_ICCV, deng2023nerdi}.
Diffusion posterior illumination~\cite{lyu2023diffusion} learns the distribution of environment illumination with a diffusion model for material recovery.
In contrast, our method learns the material prior for supervision and design it as conditional diffusion models of albedo and specular shading, enabling the direct supervision on material estimation.

\section{Method}
\label{sec:method}
\begin{figure*}[t]
    \vspace{0cm}
    \centering
       \includegraphics[width=1\linewidth]{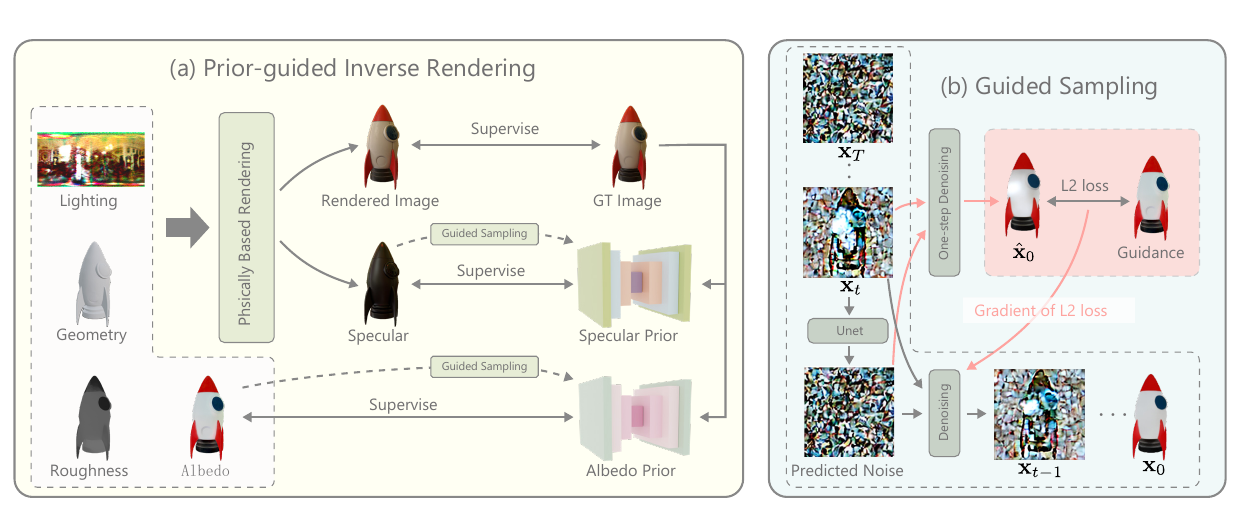}
       \caption{
           \textbf{Overview of our pipeline.}
           (a) Based on physically based rendering, our model combines lighting, geometry, roughness, and albedo into RGB and specular images, and optimizes the lighting and materials in a two-stage manner.
           In the first stage, our model is supervised by images and diffusion priors to output coarse albedo and roughness.
           Subsequently, the coarse materials are used to guide diffusion models to provide more multi-view consistent constraints.
           (b) The guided sampling first calculates the L2 loss between the guidance and one-step denoised signals, and then adds the gradient of the L1 loss to the output of the noise predictor.
        }
       \label{fig:pipeline}
\end{figure*}

Our method aims at recovering material and lighting parameters from multi-view images under a single unknown illumination. We define object material as albedo and roughness without considering the metallic attribute and use a lat-long environment map for lighting.
Due to the ill-posed nature of inverse rendering~\cite{zhang2022invrender,2022nvdiffrecmc}, ambiguities exist in the optimization process, leading to inaccurate material estimation.
We propose using data-driven prior to constrain the optimization to eliminate these ambiguities.
Inspired by the recent success of generative models~\cite{rombach2021highresolution} on image generation, we use the conditional diffusion model to learn the prior of object materials.

Fig.~\ref{fig:pipeline} presents the overview of our pipeline.
In this section, we first introduce our rendering model in Sec. \ref{sec:pbr}, and then describe how to capture the material prior using the diffusion model in Sec. \ref{sec:diffusion}. Finally, Sec. \ref{sec:joint-opt} describes our coarse-to-fine optimization strategy for accurate material and lighting recovery.

\subsection{Rendering Model}
\label{sec:pbr}

We adopt the physically based rendering \cite{10.1145/15922.15902} to recover the object material and lighting from images.
For a 3D point $\mathbf{x}$, its outgoing light $L_o$ at direction $\mathbf{\omega}_o$ is defined as:
\begin{equation}
    L_o(\omega_o;\mathbf{x}) = \int_{\Omega} L_i(\omega_i)f_r(\mathbf{x}; \omega_o, \omega_i)(\omega_i \cdot \mathbf{n}) d\omega_i,
    \label{eq:pbr_eq}
\end{equation}
where $L_i(\mathbf{\omega}_i)$ is the incident lighting along the direction $\mathbf{\omega}_i$, $\mathbf{n}$ is surface normal at $\mathbf{x}$. 
We use the simplified Disney BRDF model~\cite{burley2012physically} for $f_r$, where the albedo $\mathbf{k}_d$ and roughness $\rho$ parameters are set as learnable, and other parameters are set as constant.
The detailed formulation can be found in supplementary material.

In practice, we define the lighting as an optimizable longitude-latitude environment map. 
The geometry is obtained using MonoSDF \cite{Yu2022MonoSDF} which optimizes an implicit SDF field using posed multi-view images. Then a triangulation mesh can be extracted using Marching-Cubes \cite{lorensen1987marching}.
We further use blender \cite{blender} to build a UV mapping of the extracted mesh.
The object material albedo $\mathbf{k}_d$ and roughness $\rho$ are represented as learnable images on the UV space.

The RGB image is rendered by evaluating Eq.~\eqref{eq:pbr_eq} with the environment lighting, albedo image, roughness image, and scene mesh. We solve the integral in Eq.~\eqref{eq:pbr_eq} using Monte-Carlo sampling with Multiple Importance Sampling \cite{veach1995optimally}.
Given the rendered image, previous methods \cite{zhang2022invrender, hasselgren2022nvdiffrecmc, Jin2023TensoIR} usually optimize the material and lighting parameters by minimizing the rendered and observed images.
However, these methods easily get trapped to local minima during training due to the inevitable ambiguity, which is caused by the coupling of object material and lighting.

\paragraph{Design of the material prior.}
To better decompose materials and lighting, we propose to use data-driven priors to constraint the optimization process. Ideally, we can directly apply prior distributions of albedo and roughness to narrow down the solution space and effectively alleviate ambiguities.
However, roughness, being a concealed characteristic of a material, is difficult to observe and is also hard to acquire on most large-scale synthetic datasets~\cite{objaverse}. Because these objects have a variety of material parameters but do not necessarily have a valid roughness. 

Fortunately, image rendering using the Disney BRDF model is composed of specular shading and diffuse shading regardless of the choice of material parameters \cite{burley2012physically}.
In the Disney BRDF model, the BRDF model $f_r$ is defined as:
\begin{equation}
    f_r(\mathbf{x}; \omega_o, \omega_i) = \frac{\mathbf{k}_d}{\pi} + \frac{D \cdot G \cdot F}{4(\mathbf{n\cdot \omega_o})(\mathbf{n} \cdot \omega_i)},  
\end{equation}
where $D, G, F$ mean the normal distribution, geometry attenuation, and Fresnel effect, respectively. Then, the rendering equation is defined as:
\begin{equation}
\begin{split}
\label{eq:pbr_seperated}
L_o(\omega_o; \mathbf{x}) & = \frac{\mathbf{k}_d}{\pi}\int_{\Omega} L_i(\omega_i)(\omega_i \cdot \mathbf{n}) d\omega_i \\
&+ \int_{\Omega} L_i(\omega_i)\frac{D \cdot G \cdot F}{4(\mathbf{n\cdot \omega_o})(\mathbf{n} \cdot \omega_i)}(\omega_i \cdot \mathbf{n}) d\omega_i \\
                 & = S_{\text{diff}}(\mathbf{x}, L_i, k_d) + S_{\text{spec}}(\omega_o; \mathbf{x}, \rho, L_i). \\
\end{split}
\end{equation}
The detailed explanation of the rendering equation is presented in the supplementary material.
Motivated by Eq.~\eqref{eq:pbr_seperated}, we propose to apply priors on specular shading which exists in all physically based renderings, instead of directly using priors on roughness. Specular shading is closely linked to roughness and has the advantage of being more observable.
Moreover, based on any synthetic object, we can create the ground-truth specular shading for training the prior model.

\subsection{Albedo and Specular Prior Models}
\label{sec:diffusion}

Previous methods \cite{shi2017learning,yi2020leveraging, yu2019inverserendernet,lichy_2021,li2018cgintrinsics,YuSelfRelight20} generally model the prior of object material using a deterministic prediction network, which attempts to regress the material from a single image.
In contrast, we leverage generative models to capture the distribution of albedo and specular, considering that there could be multiple solutions for the underlying object material given captured images~\cite{barron2014shape}.

The diffusion model is a probabilistic generative model that generates images from pure Gaussian noise through a progressive de-noising process \cite{ho2020denoising}. 
Each step of the denoising process from $\mathbf{x}_{t}$ to $\mathbf{x}_{t-1}$ is defined as:
\begin{equation}
\label{eq:ddpm-ssample}
    \mathbf{x}_{t-1} = \frac{1}{\sqrt{\alpha_t}}\left(\mathbf{x}_t - \frac{1-\alpha_t}{\sqrt{1-\hat{\alpha_t}}}\mathbf{\epsilon}_{\theta}(\mathbf{x}_t, t)\right) + \sigma_t \mathbf{z},
\end{equation}
where $\mathbf{\epsilon}_{\theta}$ is the learnable noise predictor, $\alpha_{t}$ and $\sigma_t$ are predefined constant values, and $\mathbf{z}$ is a random value sampled from a standard Gaussian distribution.
The noise predictor is typically implemented as a UNet \cite{ronneberger2015u}.

Given an RGB image $I$, we can model the conditional distribution of target albedo $p_\theta(\mathbf{k}_d|I)$ and specular shading $p_\theta(S_{\rm spec}|I)$ by modifying the noise predictor to accept $I$ as its conditioning.
Denote the modified noise predictor as $\mathbf{\epsilon}_{\theta}(\mathbf{x}_t, t; I)$.
Specifically, we first use the CLIP image encoder \cite{ramanishka2018learning} to extract a feature vector from the conditioning image.
Then, to inject the conditioning signal into the UNet, we utilize a transformer network \cite{NIPS2017Attention} to perform the cross-attention between the image feature vector and the intermediate feature maps of the UNet.
More details can be found in the supplementary.

The diffusion model can model complex distributions in a step-wise denoising manner, which is crucial for albedo and specular shading recovery.
Moreover, as a generative model, the diffusion model facilitates a guided sampling procedure \cite{ho2022classifier, meng2022sdedit, chung2023diffusion}, which permits us to steer the diffusion model toward producing samples that align with multi-view image observations.

\vspace{-1em}
\arx{
\paragraph{Handle high-resolution images.}
The diffusion prior models are constrained to process images of limited resolutions (such as 256 $\times$ 256 $\times$ 3 in our implementation), which restricts their ability to handle high-resolution images and leads to the loss of fine material details.
To overcome this issue, we employ a strategy of cropping the image into smaller patches with overlapping regions and leverage the Diffusion Posterior Sampling (DPS)~\cite{chung2023diffusion} to generate the detailed material $\mathbf{x}^{f}$ for each patch.
Specifically, we first downsample the high-resolution image to predict the coarse material $\mathbf{x}^{c}$ using the diffusion model.
Then, to obtain consistent material across patches, the coarse material is used as the guidance signal to generate detailed material $\mathbf{x}^f$ for each patch based on the trained diffusion model and the Diffusion Posterior Sampling:
\begin{equation}
    \mathbf{\epsilon}^f_\theta(\mathbf{x}^f_t, t; I) = \mathbf{\epsilon}_\theta(\mathbf{x}^f_t, t; I) + \gamma_c \bigtriangledown_{\mathbf{x}^f_t}|| G(\hat{\mathbf{x}}^f_t) - \mathbf{x}_{p}^{c}
    ||_2,
    \label{eq:high-res}, 
\end{equation}
where $\mathbf{\epsilon}^f$ is the predicted noise after guidance, ${x}_{p}^{c}$ is the corresponding patch of the coarse material $\mathbf{x}^{c}$, $\mathbf{x}^f_t$ is the noisy latents at timestamp $t$, $\hat{\mathbf{x}}^f_t$ is the denoised material image of $\mathbf{x}^{f}$, and $\gamma_c$ is the guidance scale. $G$ is the Gaussian blur kernel defined in DPS\cite{chung2023diffusion}.
The final material output is obtained by computing the weighted average of all overlapping patches.
We demonstrate qualitative results in Fig.~\ref{fig:show-cases}.
}

\subsection{Material Prior for Inverse Rendering}
\label{sec:joint-opt}

Based on the albedo and specular shading prior models, we design a two-stage optimization process for inverse rendering, which effectively alleviates the ill-posed problem. An overview of our method is presented in Fig.~\ref{fig:pipeline}.
We first use the image-conditioned diffusion models to regularize the inverse rendering process, which outputs an initialization of environemnt lighting, object albedo, and object roughness.
Then, the coarse material is used to guide the sampling of the diffusion model, enabling it to produce more deterministic results and better stabilize the optimization process.

\paragraph{Training with diffusion model.}
Given input multi-view images of the target object, we utilize the conditional diffusion model to predict their albedo and specular shading components, which are used to supervise the model parameters.
Due to the probabilistic nature of the diffusion model, the generated albedo and specular images are inconsistent across camera views in terms of intensity.
To overcome this problem, we apply shift and scale invariant loss \cite{Ranftl2022} to compute the difference between the predicted and generated material.

In the first stage, the training loss function is defined as:
\begin{equation}
\begin{split}
    \mathcal{L}_{\text{coarse}} = & \lambda_1|| \hat{I} - I || + \lambda_2|| \text{SSI}(\hat{\mathbf{k}}_d, \mathbf{k}^{\rm s}_d) - \mathbf{k}^{\rm s}_d || \\ 
    & + \lambda_3|| \hat{S}_{\text{spec}} - S^{\text{s}}_{\text{spec}} || + \lambda_4\mathcal{L}_{\text{smooth}} ,
\end{split}
\label{eq:loss}
\end{equation}
where $\text{SSI}(a,b)$ aligns $a$ to the scale and mean of $b$. $\hat{I}$, $\hat{k}_d$, $\hat{S}_{spec}$ are the image, albedo, and specular shading to be optimized respectively, $\mathbf{k}^{\rm s}_d$ and $S^{\text{s}}_{\text{spec}}$ are samples from our diffusion models $p_\theta(\mathbf{k}_d|I)$ and $p_\theta(S_{\rm spec}|I)$ respectively. We also use a smoothness constraint $\mathcal{L}_{\text{smooth}}$ on the optimized materials. $\lambda_1$, $\lambda_2$, $\lambda_3$, $\lambda_4$ are hyper-parameters. More details can be found in the supplementary material.

\begin{figure}[t]
    \vspace{0cm}
    \centering
       \includegraphics[width=1\linewidth]{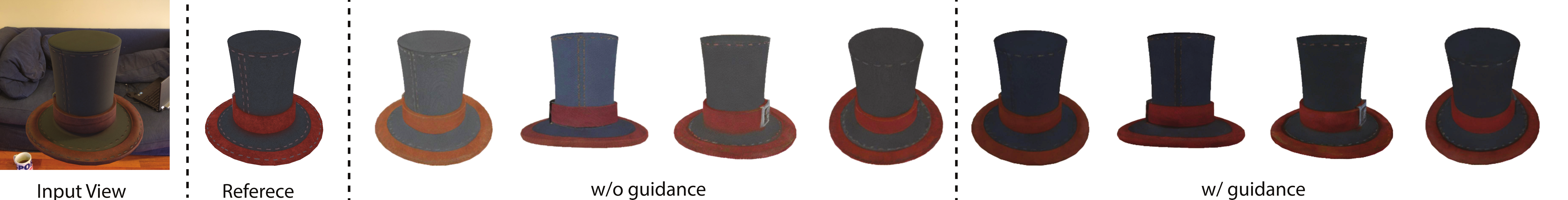}
       \caption{
           \textbf{Effect of the guided sampling.} We visualize the samples generated from the albedo prior model of different viewpoints. Without the guided sampling, the materials are inconsistent across multi-views and do not align with the material decomposition from the observed lighting.
           }
       \label{fig:guided}
\end{figure}

\paragraph{Training with guided diffusion samples.}
After the first stage, we obtain an initialization of the object material and the environment lighting.
To better guide the optimization process, we aimed to generate samples from the diffusion prior models that are consistent with multiple viewpoints, aligned with observed lighting conditions, and still follow the distribution of priors. 
To this end, we leverage the 
\arx{DPS} \cite{chung2023diffusion} to guide the sampling process, and use the estimated albedo $\mathbf{\hat{k}}_d$ and specular shading $\hat{S}_{\rm spec}$ as guiding signals to re-generate samples. As shown in Fig. ~\ref{fig:guided}, samples after guidance are multi-view consistent and fit observations under the current lighting condition.
Specifically, a guiding term $\mathbf{g}$ is added to the denoising score in Eq.~\eqref{eq:ddpm-ssample} to re-generate albedo $\tilde{\mathbf{k}}_d$ and specular $\tilde{S}_{\rm spec}$:
\arx{
\begin{equation}
    \tilde{\mathbf{\epsilon}}_\theta(\tilde{\mathbf{x}}_t, t; I) = \mathbf{\epsilon}_\theta(\tilde{\mathbf{x}}_t, t; I) + \gamma \bigtriangledown_{\tilde{\mathbf{x}}_t}|| \tilde{\mathbf{x}}^0_t - \hat{\mathbf{x}} ||_2,
    \label{eq:guided-sampling}
\end{equation}
}
\arx{
where $\tilde{\mathbf{x}}_t$ and $\tilde{\mathbf{x}}^0_t$ is the noisy latents and the denoised material image of $\tilde{\mathbf{k}}_{d}$ or $\tilde{S}_{\rm spec}$ on step $t$, respectively. $\hat{\mathbf{x}}$ is $\hat{\mathbf{k}}_d$ or $\hat{S}_{\rm spec}$, and $\gamma$ is a pre-defined scalar.
}
Finally, we use the re-generated multiview-consistent material $\tilde{\mathbf{k}}_{d}$ or $\tilde{S}_{\rm spec}$ from the DPS to supervise the lighting and materials using Eq.~\eqref{eq:loss}.
Experiments show that the second stage gives an accurate decomposition of the object material and environment lighting.

An alternative way to supervise the inverse rendering with diffusion models is the Score Distillation Sampling (SDS) \cite{poole2022dreamfusion}, which achieves consistency by finding a mode in the distribution through optimization of the material field.
However, our experiments find that the SDS guidance usually produces blurry results.
A possible reason is that the SDS guidance relies on noise prediction that has a high variance, which leads to unstable convergence as discussed in \cite{wang2023prolificdreamer}.
In constrast, our pipeline uses guided image samples that are consistent with low variance.

\section{Experiments}
\label{sec:exps}

\subsection{Implementation details}

\PAR{Prior Model Training.}
We create albedo and specular training data based on Objaverse dataset~\cite{objaverse}. We randomly select 350k objects, excluding those with pure black or white albedos, and gather 180 HDR environment maps from ~\cite{polyheaven} for rendering.
Each object is rendered under ten random poses with one random HDR environment map each. 
We use a diffusion model fine-tuned from stable-diffusion~\cite{rombach2021highresolution} for the image variation task~\cite{imgaevar} as a starting point and continue to train for 100k iterations. The training process for each prior model takes around 7 days on 8 NVIDIA A100 (80GB) GPUS.

\PAR{Inverse Rendering.}  
The resolutions for texture maps and environment are $2048 \times 2048$ and $512 \times 256$, respectively.
We use Nvdiffrast\cite{Laine2020diffrast} for rendering images with 144 samples per pixel. We train the first stage of our method for 20k iterations and the second stage for 15k iterations. The setting of hyperparameters is presented in the supplement materials. The inverse rendering for each scene takes around 1.5 hours on a single NVIDIA RTX A6000 GPU.

\subsection{Evulation}
\paragraph{Dataset.}
We collect 4 CAD models with a variety of object shapes and materials.
Each object is rendered with 120 images with 100 for training and 20 images for testing.
The training views include the posed rgb images, and their corresponding masks from Blender Cycles \cite{blender}. For evaluation, we also render 20 novel test views under original lights and novel lights along with their ground truth BRDFs.

\PAR{Baselines.}
We compare our methods with the following two types of baseline methods.
(1) \textbf{Optimization-based methods} ~\cite{2022invrender, Jin2023TensoIR, munkberg2022extracting, zhang2022invrender, zhang2023neilf++}.
InvRender \cite{2022invrender} and TensoIR \cite{Jin2023TensoIR} both models indirect illumination with MLPs and uses Spherical Gaussian(SG) as environment light. InvRender \cite{2022invrender} uses MLP representations while TensoIR is inspried by tensorial representation \cite{Chen2022ECCV}. NeILF++ \cite{zhang2023neilf++} models an incident light field that includes both environemnt light and indirect light with an MLP constrained by the outgoing radiance field. NvDiffRec \cite{Munkberg_2022_CVPR}, and NvDiffrecMC \cite{2022nvdiffrecmc} extracted explicit topology from neural SDF and directly optimize the 3D mesh with its PBR materils. Their illumination is expressed as an HDR environment map, which can be optimized by NviDiffRec's \cite{munkberg2022extracting} split sum environment lighting and NviDiffRec's Monte Carlo ray tracing.
(2) \textbf{Data-driven methods}~\cite{Liu2020unsupervised, yi2023weaklysupervised}.
We keep object and lighting models as the same and just replace diffusion priors with single-view intrinsic image decomposition methods for comparison.
USI3D \cite{Liu2020unsupervised} models the decomposition task as domain transfer from the image domain to the albedo or shading domain using generative models. We adopt only the predicted albedo to supervise the optimization because their rendering model is not physically based. Yi. el \cite{yi2023weaklysupervised} propose a network to predict and remove specular shading and subsequently recover the albedo. We use the predicted specular shading and albedo as the priors.

\PAR{Evaluation protocol.}
We analyze our method and the baselines in the following way. On the 4 synthetic datasets, we compare their reconstruction of albedo and roughness quantitatively with the ground truths. With the estimated albedo and lights, we also render the novel view images to further illustrate the quality of the predicted materials and lights. Given the environment map from our novel light dataset, we also make a comparison on the relighted images. Because Invrender \cite{zhang2022invrender} and TensoIR \cite{Jin2023TensoIR} parameterize environment maps with SGs, we first optimize an SG with 128 lobes through Stochastic Gradient Decent optimization. For NvDiffRec \cite{munkberg2022extracting}, NvDiffRecMC \cite{2022nvdiffrecmc}, and NeILF++\cite{zhang2023neilf++}, we render the relighted image with Blender \cite{blender} using our extracted topology and material texture maps with SPP=4096 and 4 bounces of lights. Due to the inevitable scale ambiguity between lighting and albedo, previous methods \cite{nerfactor} align the predicted albedo to the ground truth before relighting and evaluation. Each channel is separately aligned to the ground truth. However, this would cause the incorrectly baked shading in the albedo to be removed during alignment. To this end, we align the average RGB of prediction to ground truth for a more faithful evaluation. Because geometry affects the inverse rendering performance greatly, to make sure the topology for NVDiffRec and NvdiffRecMC is well initialized, we make another set of experiments for NvDiffRec and NvDiffRecMC with MonoSDF \cite{Yu2022MonoSDF} reconstructed geometry given and as a fixed topology along the training and evaluation process.

\PAR{Metrics.}
 We use Peak Signal-to-Noise Ratio (PSNR), Structural Similarity Index Measure (SSIM), and Learned Perceptual Image Patch Similarity (LPIPS)\cite{zhang2018perceptual} as metrics to evaluate the image quality of the aligned albedo, relighting images, and novel view synthesis. Roughness is evaluated using Mean Squared Error(MSE).

\begin{table*}[t!]
\centering
\caption{
\textbf{Quantitative results on the synthetic dataset.}
``NvDiffrec*'' and ``NvDiffrecMC*'' mean these methods are with our reconstructed mesh.
``Ours w/o guided'' means our method without the guided sampling strategy.
TensorIR and NvDiffrec have strong capabilities of fitting observed images, but they struggle to recover correct object materials.
Replacing our diffusion priors with other priors~\cite{yi2023weaklysupervised,Liu2020unsupervised} leads to the degradation of the performance on material recovery.
}

\resizebox{1.0\textwidth}{!}{
\setlength{\tabcolsep}{2pt} %
\renewcommand{\arraystretch}{1.2} %
\begin{tabular}{ccccccccccccccccccc} 
\toprule
                                                                        &  & \multicolumn{3}{c}{Albedo} &  & \multicolumn{3}{c}{Aligned Albedo} &  & \multicolumn{3}{c}{View Synthesis} &  & \multicolumn{3}{c}{Relighting} &  & Roughness  \\ 
\cline{3-5}\cline{7-9}\cline{11-13}\cline{15-17}\cline{19-19}
Method                                                                  &  & PSNR$\uparrow$ & SSIM$\uparrow$ & LPIPS$\downarrow$        &  & PSNR$\uparrow$ & SSIM$\uparrow$ & LPIPS$\downarrow$                &  & PSNR$\uparrow$ & SSIM$\uparrow$ & LPIPS$\downarrow$                &  & PSNR$\uparrow$ & SSIM$\uparrow$ & LPIPS$\downarrow$            &  & MSE$\downarrow$        \\ 
\midrule
InvRender \cite{2022invrender}     &  
& 17.8181    & 0.9247    & 0.0645           & 
& 23.7592    & 0.9491    & 0.0559    &
& 27.3051    & 0.9673    & 0.0386                    &  
& 24.3632    & 0.9621  &  0.0515   & 
& 0.0420          \\
TensoIR  \cite{Jin2023TensoIR}    &  
& 20.2972   &  0.9527  &   0.0631        &
 
& 24.9368   & \textbf{0.9564}   & 0.0614               & 

&   \textbf{33.1234}   &  0.9803   &0.0264             &  
& 21.3473   &  0.9543    &  0.0514              & 
& 0.0456          \\
NvDiffrec                  \cite{munkberg2022extracting}                                             &  
& 16.5467   &  0.9240    &    0.0706  &
& 21.5304    & 0.8972    &  0.1241                &  
& 32.5778   &   \textbf{0.9824}   &     \textbf{0.0190}   & 
& 21.9939  & 0.9550 &        0.0734             &  
& 0.0777          \\
NvDiffrec* \cite{munkberg2022extracting}   &  
&   14.5204   &  0.9021   &    0.0835          & 
&  20.2280   &   0.9204   &     0.0659                 & 
&   31.0336   &   0.9784   &     0.0218                 &  
&   21.2452  &  0.9356 &        0.0638        &  
&     0.0841       \\
NvDiffrecMC              \cite{2022nvdiffrecmc}                                             &  
& 13.0495    & 0.8815    & 0.1068           &  
& 21.6770    & 0.9406    &  0.0699                 &  
& 30.1083    & 0.9674    & 0.0440                   &  
& 22.5337    & 0.9486    & 0.0797              &  
& 0.1046        \\
NvDiffrecMC*\cite{2022nvdiffrecmc} & 
& 11.1353    & 0.8722    & 0.1176            & 
& 19.8325    & 0.9305    & 0.0805                & 
& 28.1634    & 0.9560    & 0.0440                    & 
& 21.0380    & 0.9332    & 0.0822               &  
&  0.1411          \\ 
NEILF++*\cite{zhang2023neilf++} & 
& 13.8336    & 0.9172    & 0.0915            & 
& 18.4543    & 0.9366    & 0.0831                & 
& 29.3527    & 0.9775    & 0.0276                    & 
& 20.4878    & 0.9629    & 0.0590               &  
&  0.0855          \\ 
\midrule
Yi \etal  \cite{yi2023weaklysupervised}                                                   
 & & 13.2029 & 0.8934 & 0.0833 &  & 18.8428 & 0.9133 & 0.0731 & & 26.5091 & 0.9669 & 0.0393 & & 21.156 & 0.9562 & 0.0507 & & 0.1524

    \\
USI3D  \cite{Liu2020unsupervised}                                                  
 & & 11.6713 & 0.8872 & 0.1051 &  & 18.9296 & 0.9344 & 0.0772 & & 26.9653 & 0.9666 & 0.0480 & & 20.2721 & 0.9587 & 0.0612 & & 0.1241

    \\
\midrule
Ours                                                                
 & & \textbf{24.6766} & \textbf{0.9562} & \textbf{0.0503} &  & \textbf{26.8829} & 0.9532 & \textbf{0.0378} & & 30.2405 & 0.9741 & 0.0328 & & \textbf{28.9721} & \textbf{0.9757} & 0.0289 & & \textbf{0.0105}
    \\

Ours w/o spec                                                   
 & & 17.2126 & 0.9316 & 0.0567 &  & 25.7389 & 0.9491 & 0.0444 & & 29.4245 & 0.9728 & 0.0352 & & 26.4743 & 0.9729 & 0.0363 & & 0.0758
 \\
 Ours w/o albedo                                                   
 & & 22.7111 & 0.9434 & 0.0571 &  & 24.3462 & 0.9381 & 0.0488 & & 30.5998 & 0.9751 & 0.0261 & & 27.2851 & 0.9743 & \textbf{0.0258} & & 0.0118
 \\
Ours w/o guided                                                   
 & & 23.3507 & 0.9303 & 0.0652 &  & 24.6112 & 0.9208 & 0.0641 & & 29.7937 & 0.9688 & 0.0339 & & 27.7390 & 0.9683 & 0.0361 & & 0.0168
 \\
\bottomrule
\end{tabular}
}
\label{tab:syn-results}
    
\end{table*}

\begin{figure*}[!t]
    \centering
       \includegraphics[width=1\linewidth]{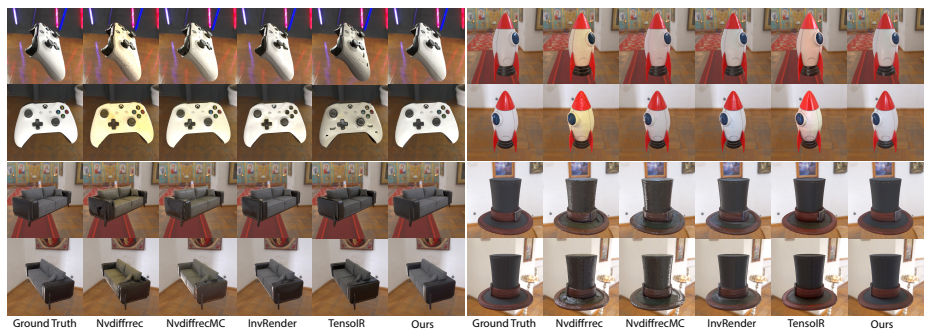}
       \caption{
           \textbf{Qualitative comparison in terms of relighting} on the synthetic dataset.
           Zoom in for details.
        }
       \label{fig:syn-relighting}
\end{figure*}

\begin{figure}[!t]
    \centering
       \includegraphics[width=1\linewidth]{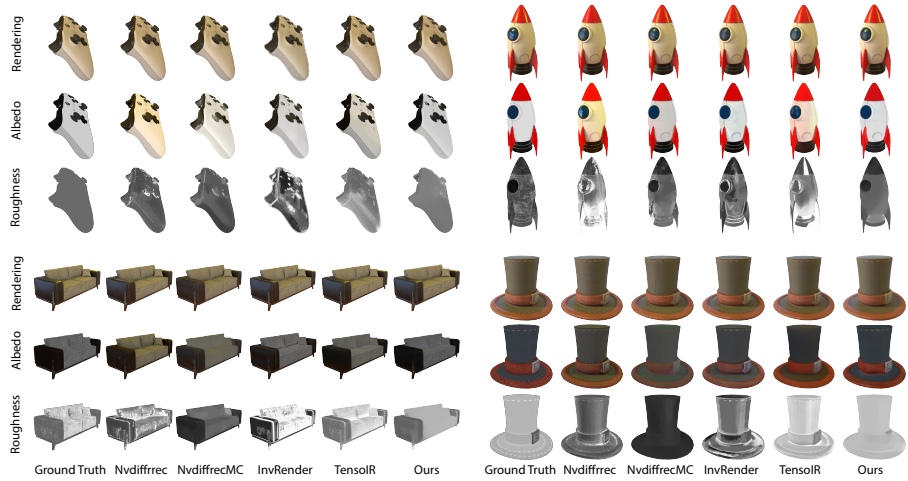}
       \caption{
           \textbf{Quantitative comparison with baselines on the synthetic dataset.}
           Our method estimates more accurate albedo and roughness than other methods.
       }
       \label{fig:syn-materials}
\end{figure}

\PAR{Comparison with optimization-based methods.}
Tab.~\ref{tab:syn-results} quantitatively compares our method with recent inverse rendering methods on the synthetic dataset.
We significantly outperforms baseline methods in terms of recovered albedo, roughness, and relighting, demonstrating the effectiveness of the proposed material prior.
Although TensoIR~\cite{Jin2023TensoIR} and NvDiffRec~\cite{Munkberg_2022_CVPR} shows good results on view synthesis under the original environment lighting, it does not accurately estimate the underlying materials, which will be further shown in the qualitative results.
Tab.~\ref{tab:syn-results} also shows that we further improve the performance of material recovery by guiding the diffusion model with estimated albedo and specular.

\begin{figure*}[!t]
    \centering
       \includegraphics[width=1.0\linewidth]{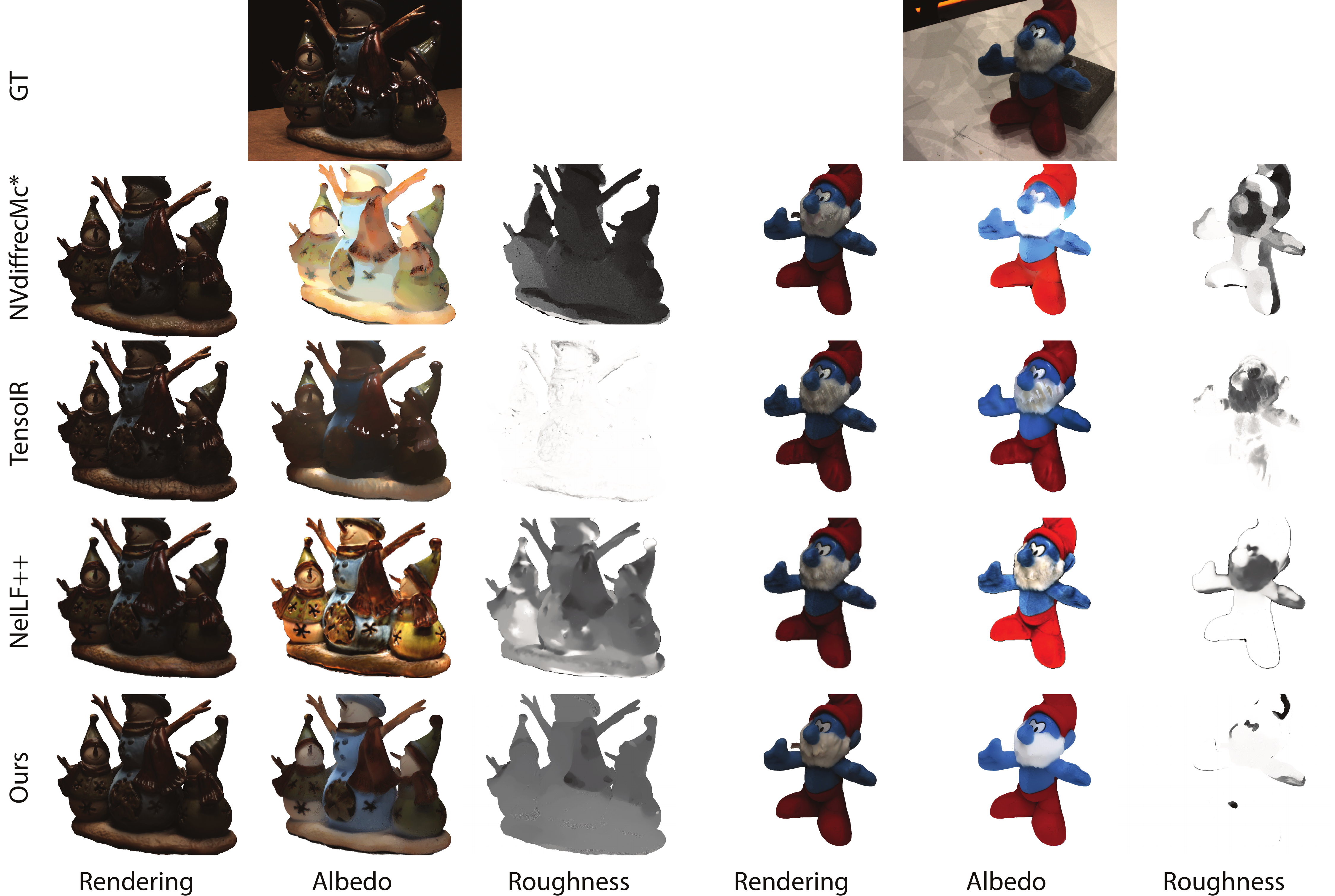}
       \caption{
           \textbf{Qualitative results on DTU.} Our method recovers reasonable materials on real datasets. Baseline models can not disentangle illumination and materials which results in the baked-in effect on materials estimation.
           }
       \label{fig:dtu-material}
\end{figure*}

\begin{figure*}[!t]
    \centering
       \includegraphics[width=1\linewidth]{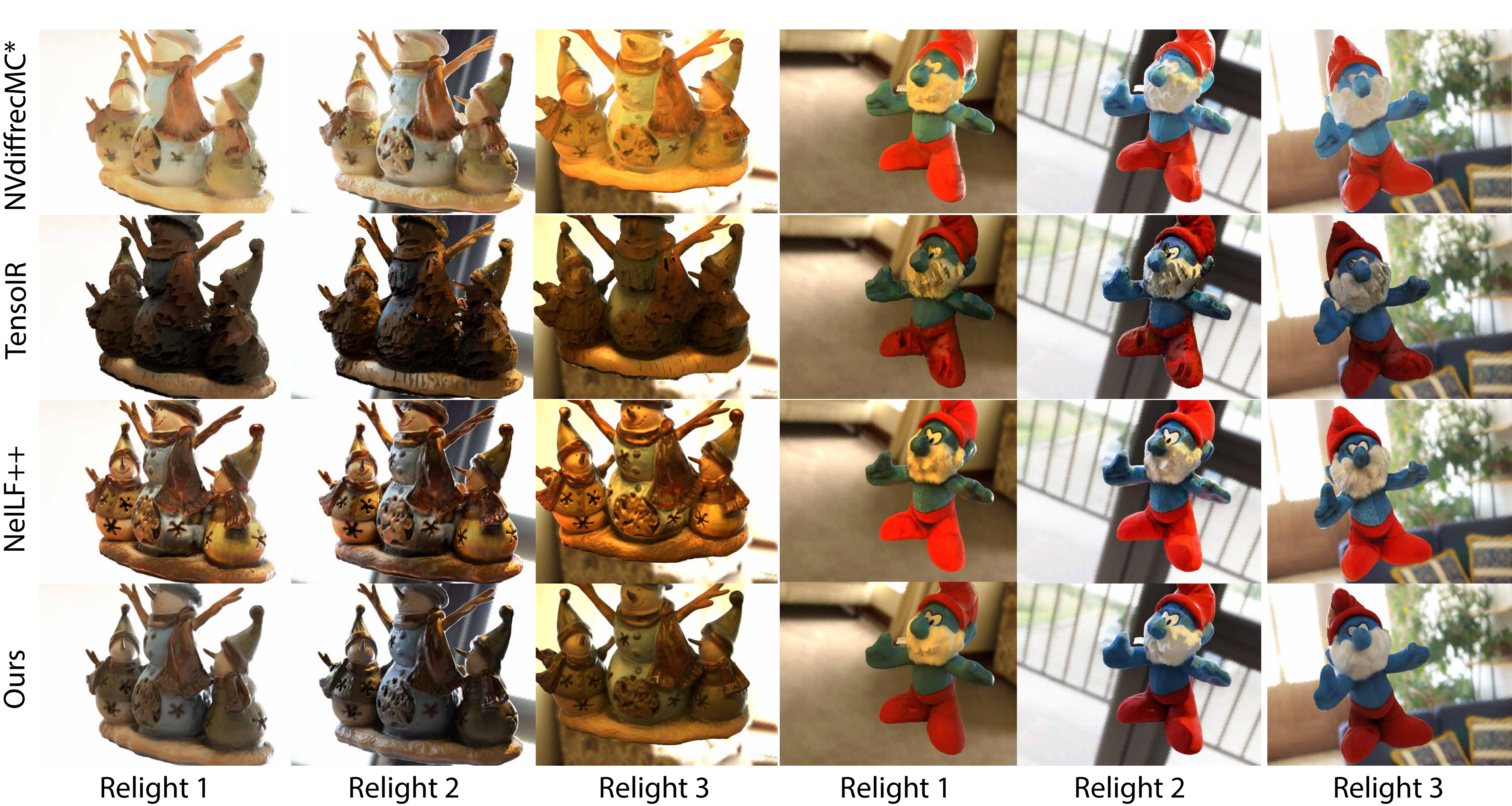}
       \caption{
           \textbf{Qualitative results of relighting on DTU.} 
           }
       \label{fig:dtu-relight}
\end{figure*}

\begin{figure}[t]
    \vspace{0cm}
    \centering
       \includegraphics[width=1\linewidth]{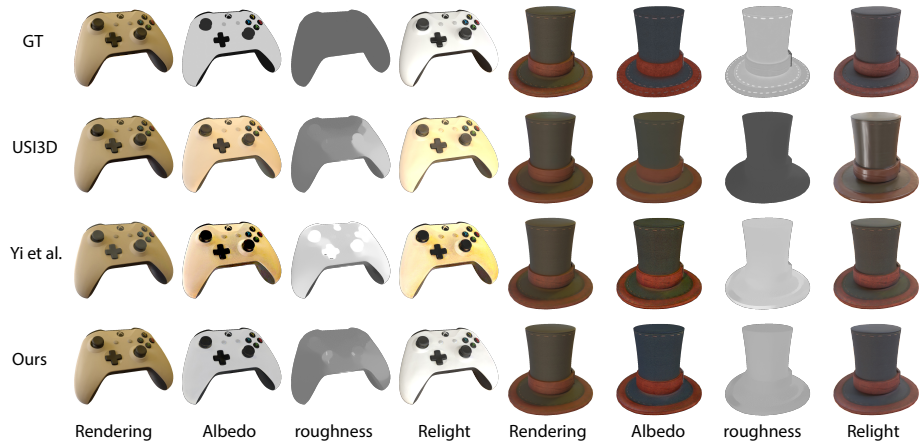}
       \caption{
           \textbf{Comparison with other data-driven priors.}
           }
       \label{fig:data-driven}
\end{figure}

Fig.~\ref{fig:syn-relighting} and Fig.~\ref{fig:syn-materials} present the qualitative results on relighting and estimated materials.
Fig.~\ref{fig:dtu-material} and Fig.~\ref{fig:dtu-relight} demonstrate the performance of our method on the real-world data.
More results on real data are presented in the supplementary material.
NvDiffrec and NeILF++ does not disentangle the albedo and light well and show strong light color baked in albedo, as visualized in Fig.~\ref{fig:syn-materials} and Fig.~\ref{fig:dtu-material}.
This failure case is further demonstrated in Fig.~\ref{fig:syn-relighting} and Fig.~\ref{fig:dtu-relight}.
Such baked-in issue can be partially resolved by the InvRender's sparsity constraint.
However, this sparsity constraint may result in lost of details, as shown in Fig.~\ref{fig:syn-materials}.
With our diffusion priors learned from the large-scale dataset, our method can disentangle materials and lighting in a reasonable way.

\PAR{Comparison with other data-driven priors.}
Tab.~\ref{tab:syn-results} presents the comparison with data-driven priors~\cite{yi2023weaklysupervised, Liu2020unsupervised}.
Although these methods are able to prevent shadows from being baked into albedo, they still struggle to separate shading from the albedo of objects under complex lighting conditions, as shown in qualitative results in Fig.~\ref{fig:data-driven}. The incorrect priors will mislead the optimization process and cause large errors in the estimated materials. In contrast, our prior models have a better capacity to decouple shading from materials.

\begin{figure}[t!]
    \vspace{0cm}
    \centering
       \includegraphics[width=1\linewidth]{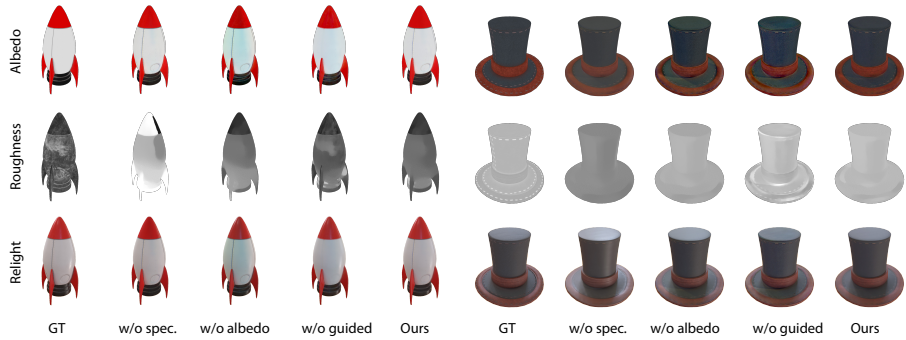}
       \caption{
           \textbf{Ablation studies} on two synthetic objects. 
           }
       \label{fig:ablation}
\end{figure}

\PAR{Ablation Study.}
We conduct ablation studies on our diffusion priors and the guided diffusion for fine-level inverse rendering. Quantitative and qualitative results are shown in Tab.~\ref{tab:syn-results} and Fig.~\ref{fig:ablation}, respectively.

\begin{figure}[ht!]
    \vspace{0cm}
    \centering
       \includegraphics[width=0.97\linewidth]{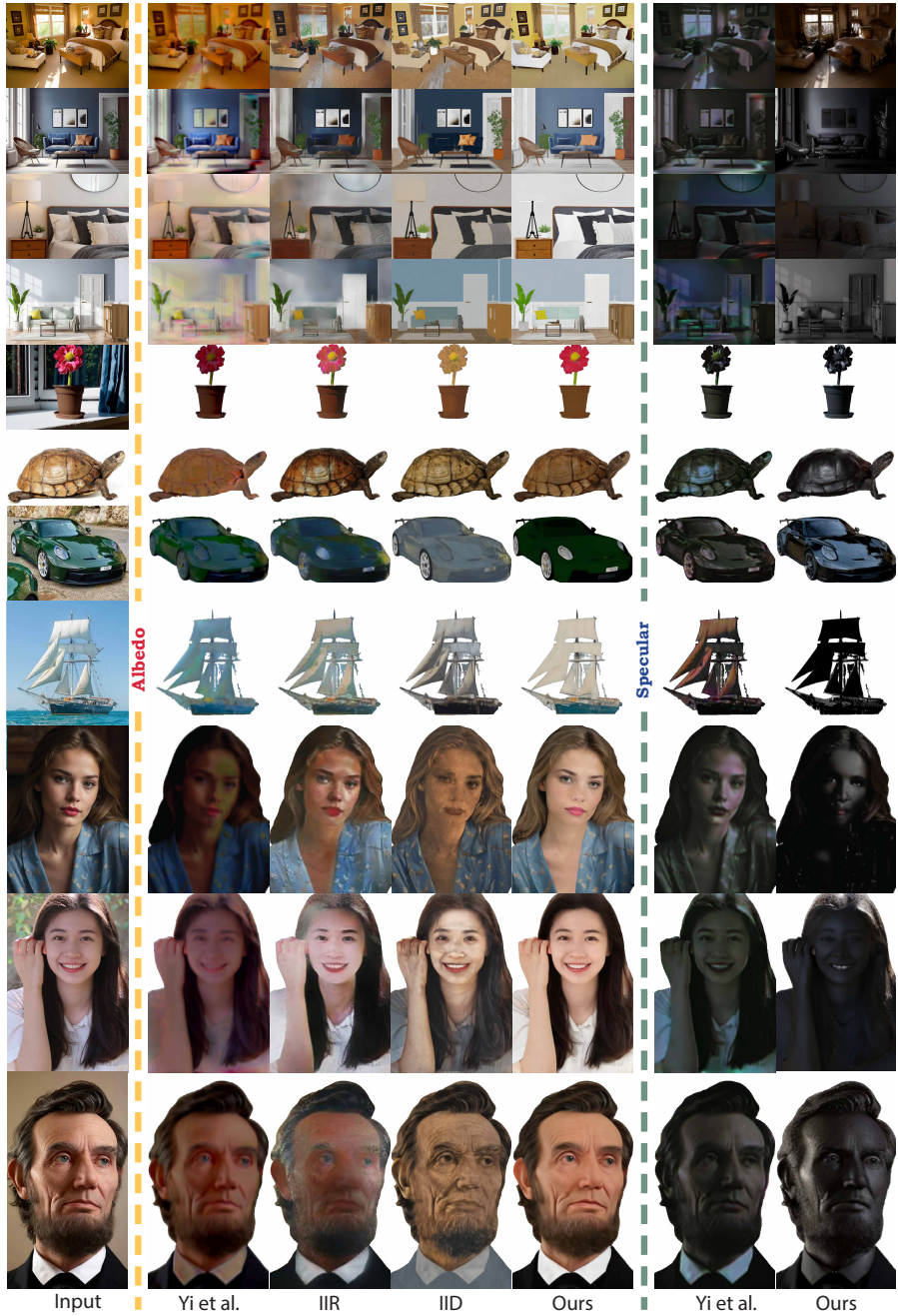}
       \caption{
           \textbf{Qualitative results on intrinsic image decomposition.}
           }
       \label{fig:show-cases}
\end{figure}

"w/o spec." indicates training without the specular shading prior model. This would result in recovering incorrect lighting and roughness, as the diffuse shading and specular shading cannot be correctly separated by relying solely on the albedo prior. As depicted in Fig. \ref{fig:ablation}, the "hat"'s roughness prediction is lower and the "ship"'s is higher in "w/o Spec." than ground truth, which resulting in inaccurate relighting outcomes. In "w/o albedo", the inverse rendering is conducted without albedo prior. The recovered albedo has baked-in shading, as shown in  Fig. \ref{fig:ablation} where the two objects baked shading into the albedo causing incorrect relighting. This indicates that both priors are crucial for accurate lighting and material decomposition.

"w/o guided" is the result without the guided diffusion for fine-level optimization. Due to the lack of multi-view consistency of diffusion prior, the optimized materials are noisy and have baked-in shading, as shown in Fig.~\ref{fig:ablation}. This hurts the relighting quality, as shown in qualitative results in Tab.~\ref{tab:syn-results}. Therefore, the training process using the guided diffusion model is essential for a smooth and accurate material recovery.

\PAR{Modeling indirect illumination and metallic materials.} Following the setting of InvRender \cite{2022invrender}, we extend our method to handle indirect lighting. Inspired by \cite{fipt2023}, we precompute shading of indirect lighting for each image by tracing the trained SDF and Color field. We also include metallic texture as an optimization target.  More details and results can be found in the supplementary materials.

\PAR{Single-view intrinsic image decomposition results.} We conduct experiments on challenging in-the-wild images spanning diverse domains, including scenes, objects, and humans. 
We compare our method's albedo prediction with an object-level method Yi \etal ~\cite{yi2023weaklysupervised}, and two scene-level methods, IIR~\cite{zhu2022learning} and IID~\cite{kocsis2023iid}. 
The specular shading prediction is compared against Yi \etal ~\cite{yi2023weaklysupervised}. 
Qualitative results are shown in Fig.~\ref{fig:show-cases}. 
Yi \etal~\cite{yi2023weaklysupervised} and IIR~\cite{zhu2022learning} cannot disentangle shading from materials, while our method recovers shading-free albedos. 
IID~\cite{kocsis2023iid} distorts the true color and introduces artifacts on the albedo, particularly noticeable on human faces. In contrast, our methods maintain the original color and preserve finer details.
Our method also recovers better specular shading, whereas the baseline mostly failed.

\section{Conclusion}

We proposed to learn the material prior for inverse rendering under an unknown static lighting condition.
To this end, the material prior is designed as conditional diffusion models of albedo and specular shading components, according to the general rendering equation.
In addition, a two-stage training scheme is developed for robustly regularizing the inverse rendering process with conditional diffusion models, which uses coarse materials from the first stage training to guide diffusion models to provide multi-view consistent constraints.
The proposed model significantly outperforms baseline methods on real-world and synthetic data.

\paragraph{Limitations.}
Although we achieved impressive results on material recovery, our approach still has some limitations.
First, we do not consider the inverse rendering of transparent objects.
A solution is revising the geometry representation to a neural field of transparency and jointly optimizing it with object materials and environment lighting.
Second, the performance of our model is limited by the accuracy of the reconstructed geometry.
It is interesting to exploit diffusion models to better learn the geometry prior for 3D reconstruction.

\newpage
\bibliographystyle{splncs04}
\bibliography{main}

\begin{thebibliography}{10}
\providecommand{\url}[1]{\texttt{#1}}
\providecommand{\urlprefix}{URL }
\providecommand{\doi}[1]{https://doi.org/#1}

\bibitem{polyheaven}
Poly haven: The public 3d asset library. \url{https://polyhaven.com/}

\bibitem{imgaevar}
Stable diffusion image variations. \url{https://huggingface.co/spaces/lambdalabs/stable-diffusion-image-variations}

\bibitem{anciukevivcius2023renderdiffusion}
Anciukevi{\v{c}}ius, T., Xu, Z., Fisher, M., Henderson, P., Bilen, H., Mitra, N.J., Guerrero, P.: Renderdiffusion: Image diffusion for 3d reconstruction, inpainting and generation. In: CVPR (2023)

\bibitem{barron2014shape}
Barron, J.T., Malik, J.: Shape, illumination, and reflectance from shading. TPAMI  \textbf{37}(8),  1670--1687 (2014)

\bibitem{bi2021avatar}
Bi, S., Lombardi, S., Saito, S., Simon, T., Wei, S.E., McPhail, K., Ramamoorthi, R., Sheikh, Y., Saragih, J.: Deep relightable appearance models for animatable faces. ACM Trans. Graph. (Proc. SIGGRAPH)  \textbf{40}(4) (2021)

\bibitem{blender}
Blender: Blender, \url{https://www.blender.org/}

\bibitem{boss2021nerd}
Boss, M., Braun, R., Jampani, V., Barron, J.T., Liu, C., Lensch, H.P.: Nerd: Neural reflectance decomposition from image collections. In: IEEE International Conference on Computer Vision (ICCV) (2021)

\bibitem{burley2012physically}
Burley, B., Studios, W.D.A.: Physically-based shading at disney. In: Acm Siggraph. vol.~2012, pp.~1--7. vol. 2012 (2012)

\bibitem{shapenet}
Chang, A.X., Funkhouser, T.A., Guibas, L.J., Hanrahan, P., Huang, Q., Li, Z., Savarese, S., Savva, M., Song, S., Su, H., Xiao, J., Yi, L., Yu, F.: Shapenet: An information-rich 3d model repository. CoRR  \textbf{abs/1512.03012} (2015), \url{http://arxiv.org/abs/1512.03012}

\bibitem{Chen2022ECCV}
Chen, A., Xu, Z., Geiger, A., Yu, J., Su, H.: Tensorf: Tensorial radiance fields. In: European Conference on Computer Vision (ECCV) (2022)

\bibitem{chen2023single}
Chen, H., Gu, J., Chen, A., Tian, W., Tu, Z., Liu, L., Su, H.: Single-stage diffusion nerf: A unified approach to 3d generation and reconstruction. In: ICCV (2023)

\bibitem{cheng2023sdfusion}
Cheng, Y.C., Lee, H.Y., Tulyakov, S., Schwing, A.G., Gui, L.Y.: Sdfusion: Multimodal 3d shape completion, reconstruction, and generation. In: CVPR (2023)

\bibitem{chung2023diffusion}
Chung, H., Kim, J., Mccann, M.T., Klasky, M.L., Ye, J.C.: Diffusion posterior sampling for general noisy inverse problems. In: The Eleventh International Conference on Learning Representations (2023), \url{https://openreview.net/forum?id=OnD9zGAGT0k}

\bibitem{chung2022come}
Chung, H., Sim, B., Ye, J.C.: Come-closer-diffuse-faster: Accelerating conditional diffusion models for inverse problems through stochastic contraction. In: Proceedings of the IEEE/CVF Conference on Computer Vision and Pattern Recognition. pp. 12413--12422 (2022)

\bibitem{croitoru2023diffusion}
Croitoru, F.A., Hondru, V., Ionescu, R.T., Shah, M.: Diffusion models in vision: A survey. T-PAMI  (2023)

\bibitem{10.1145/344779.344855}
Debevec, P., Hawkins, T., Tchou, C., Duiker, H.P., Sarokin, W., Sagar, M.: Acquiring the reflectance field of a human face. p. 145–156. SIGGRAPH '00, ACM Press/Addison-Wesley Publishing Co., USA (2000). \doi{10.1145/344779.344855}, \url{https://doi.org/10.1145/344779.344855}

\bibitem{debevec2000acquiring}
Debevec, P., Hawkins, T., Tchou, C., Duiker, H.P., Sarokin, W., Sagar, M.: Acquiring the reflectance field of a human face. In: Proceedings of the 27th annual conference on Computer graphics and interactive techniques. pp. 145--156 (2000)

\bibitem{objaverse}
Deitke, M., Schwenk, D., Salvador, J., Weihs, L., Michel, O., VanderBilt, E., Schmidt, L., Ehsani, K., Kembhavi, A., Farhadi, A.: Objaverse: A universe of annotated 3d objects. In: Proceedings of the IEEE/CVF Conference on Computer Vision and Pattern Recognition. pp. 13142--13153 (2023)

\bibitem{deng2023nerdi}
Deng, C., Jiang, C., Qi, C.R., Yan, X., Zhou, Y., Guibas, L., Anguelov, D., et~al.: Nerdi: Single-view nerf synthesis with language-guided diffusion as general image priors. In: Proceedings of the IEEE/CVF Conference on Computer Vision and Pattern Recognition. pp. 20637--20647 (2023)

\bibitem{10.1145/3355089.3356571}
Guo, K., Lincoln, P., Davidson, P., Busch, J., Yu, X., Whalen, M., Harvey, G., Orts-Escolano, S., Pandey, R., Dourgarian, J., Tang, D., Tkach, A., Kowdle, A., Cooper, E., Dou, M., Fanello, S., Fyffe, G., Rhemann, C., Taylor, J., Debevec, P., Izadi, S.: The relightables: Volumetric performance capture of humans with realistic relighting. ACM Trans. Graph.  \textbf{38}(6) (nov 2019). \doi{10.1145/3355089.3356571}, \url{https://doi.org/10.1145/3355089.3356571}

\bibitem{hasselgren2022nvdiffrecmc}
Hasselgren, J., Hofmann, N., Munkberg, J.: {Shape, Light, and Material Decomposition from Images using Monte Carlo Rendering and Denoising}. NeurIPS  (2022)

\bibitem{2022nvdiffrecmc}
Hasselgren, J., Hofmann, N., Munkberg, J.: {Shape, Light, and Material Decomposition from Images using Monte Carlo Rendering and Denoising}. arXiv:2206.03380  (2022)

\bibitem{ho2020denoising}
Ho, J., Jain, A., Abbeel, P.: Denoising diffusion probabilistic models. In: NeurIPS (2020)

\bibitem{ho2022classifier}
Ho, J., Salimans, T.: Classifier-free diffusion guidance. arXiv preprint arXiv:2207.12598  (2022)

\bibitem{horita2022structure}
Horita, D., Yang, J., Chen, D., Koyama, Y., Aizawa, K.: A structure-guided diffusion model for large-hole diverse image completion. arXiv preprint arXiv:2211.10437  (2022)

\bibitem{Jin2023TensoIR}
Jin, H., Liu, I., Xu, P., Zhang, X., Han, S., Bi, S., Zhou, X., Xu, Z., Su, H.: Tensoir: Tensorial inverse rendering. In: CVPR (2023)

\bibitem{10.1145/15922.15902}
Kajiya, J.T.: The rendering equation. In: Proceedings of the 13th Annual Conference on Computer Graphics and Interactive Techniques. p. 143–150. SIGGRAPH '86, Association for Computing Machinery, New York, NY, USA (1986). \doi{10.1145/15922.15902}, \url{https://doi.org/10.1145/15922.15902}

\bibitem{kawar2022denoising}
Kawar, B., Elad, M., Ermon, S., Song, J.: Denoising diffusion restoration models. arXiv preprint arXiv:2201.11793  (2022)

\bibitem{kerbl3Dgaussians}
Kerbl, B., Kopanas, G., Leimk{\"u}hler, T., Drettakis, G.: 3d gaussian splatting for real-time radiance field rendering. ACM Transactions on Graphics  \textbf{42}(4) (July 2023), \url{https://repo-sam.inria.fr/fungraph/3d-gaussian-splatting/}

\bibitem{kocsis2023iid}
Kocsis, P., Sitzmann, V., Nie{\ss}ner, M.: Intrinsic image diffusion for single-view material estimation. In: arxiv (2023)

\bibitem{Laine2020diffrast}
Laine, S., Hellsten, J., Karras, T., Seol, Y., Lehtinen, J., Aila, T.: Modular primitives for high-performance differentiable rendering. ACM Transactions on Graphics  \textbf{39}(6) (2020)

\bibitem{Levoy2000TheDM}
Levoy, M., Pulli, K., Curless, B., Rusinkiewicz, S., Koller, D., Pereira, L., Ginzton, M., Anderson, S.E., Davis, J., Ginsberg, J., Shade, J., Fulk, D.: The digital michelangelo project: 3d scanning of large statues. Proceedings of the 27th annual conference on Computer graphics and interactive techniques  (2000), \url{https://api.semanticscholar.org/CorpusID:1546062}

\bibitem{Li2018ECCV}
Li, Z., Snavely, N.: Cgintrinsics: Better intrinsic image decomposition through physically-based rendering. In: Proceedings of the European Conference on Computer Vision (ECCV) (September 2018)

\bibitem{li2018cgintrinsics}
Li, Z., Snavely, N.: Cgintrinsics: Better intrinsic image decomposition through physically-based rendering. In: ECCV (2018)

\bibitem{Li2018sigasia}
Li, Z., Xu, Z., Ramamoorthi, R., Sunkavalli, K., Chandraker, M.: Learning to reconstruct shape and spatially-varying reflectance from a single image. In: ACM Transactions on Graphics. vol.~37, pp. 1--11 (12 2018). \doi{10.1145/3272127.3275055}

\bibitem{li2018learning}
Li, Z., Xu, Z., Ramamoorthi, R., Sunkavalli, K., Chandraker, M.: Learning to reconstruct shape and spatially-varying reflectance from a single image. ToG  \textbf{37}(6),  1--11 (2018)

\bibitem{li2021openrooms}
Li, Z., Yu, T.W., Sang, S., Wang, S., Song, M., Liu, Y., Yeh, Y.Y., Zhu, R., Gundavarapu, N., Shi, J., Bi, S., Xu, Z., Yu, H.X., Sunkavalli, K., Hašan, M., Ramamoorthi, R., Chandraker, M.: Openrooms: An end-to-end open framework for photorealistic indoor scene datasets (2021)

\bibitem{liang2023gs}
Liang, Z., Zhang, Q., Feng, Y., Shan, Y., Jia, K.: Gs-ir: 3d gaussian splatting for inverse rendering. arXiv preprint arXiv:2311.16473  (2023)

\bibitem{lichy2021}
Lichy, D., Wu, J., Sengupta, S., Jacobs, D.W.: Shape and material capture at home. CoRR  \textbf{abs/2104.06397} (2021), \url{https://arxiv.org/abs/2104.06397}

\bibitem{lichy_2021}
Lichy, D., Wu, J., Sengupta, S., Jacobs, D.W.: Shape and material capture at home. In: CVPR (2021)

\bibitem{liu2023syncdreamer}
Liu, Y., Lin, C., Zeng, Z., Long, X., Liu, L., Komura, T., Wang, W.: Syncdreamer: Generating multiview-consistent images from a single-view image. arXiv preprint arXiv:2309.03453  (2023)

\bibitem{liu2023nero}
Liu, Y., Wang, P., Lin, C., Long, X., Wang, J., Liu, L., Komura, T., Wang, W.: Nero: Neural geometry and brdf reconstruction of reflective objects from multiview images. In: SIGGRAPH (2023)

\bibitem{Liu2020unsupervised}
Liu, Y., Li, Y., You, S., Lu, F.: Unsupervised learning for intrinsic image decomposition from a single image. In: CVPR (2020)

\bibitem{lorensen1987marching}
Lorensen, W.E., Cline, H.E.: Marching cubes: A high resolution 3d surface construction algorithm. ACM siggraph computer graphics  \textbf{21}(4),  163--169 (1987)

\bibitem{lyu2023diffusion}
Lyu, L., Tewari, A., Habermann, M., Saito, S., Zollh{\"o}fer, M., Leimk{\"u}hler, T., Theobalt, C.: Diffusion posterior illumination for ambiguity-aware inverse rendering. arXiv preprint arXiv:2310.00362  (2023)

\bibitem{meng2022sdedit}
Meng, C., He, Y., Song, Y., Song, J., Wu, J., Zhu, J.Y., Ermon, S.: {SDE}dit: Guided image synthesis and editing with stochastic differential equations. In: International Conference on Learning Representations (2022)

\bibitem{mildenhall2020nerf}
Mildenhall, B., Srinivasan, P.P., Tancik, M., Barron, J.T., Ramamoorthi, R., Ng, R.: Nerf: Representing scenes as neural radiance fields for view synthesis. In: ECCV (2020)

\bibitem{munkberg2022extracting}
Munkberg, J., Hasselgren, J., Shen, T., Gao, J., Chen, W., Evans, A., M{\"u}ller, T., Fidler, S.: Extracting triangular 3d models, materials, and lighting from images. In: CVPR (2022)

\bibitem{Munkberg_2022_CVPR}
Munkberg, J., Hasselgren, J., Shen, T., Gao, J., Chen, W., Evans, A., M\"uller, T., Fidler, S.: {Extracting Triangular 3D Models, Materials, and Lighting From Images}. In: Proceedings of the IEEE/CVF Conference on Computer Vision and Pattern Recognition (CVPR). pp. 8280--8290 (June 2022)

\bibitem{MobileSVBRDF:SIGA:2018}
Nam, G., Lee, J.H., Gutierrez, D., Kim, M.H.: Practical svbrdf acquisition of 3d objects with unstructured flash photography. ACM Transactions on Graphics (Proc. SIGGRAPH Asia 2018)  \textbf{37}(6),  267:1--12 (2018). \doi{10.1145/3272127.3275017}, \url{http://dx.doi.org/10.1145/3272127.3275017}

\bibitem{poole2022dreamfusion}
Poole, B., Jain, A., Barron, J.T., Mildenhall, B.: Dreamfusion: Text-to-3d using 2d diffusion. In: ICLR (2023)

\bibitem{ramanishka2018learning}
Ramanishka, V., Das, A., Zhang, J., Saenko, K.: Learning transferable visual models from natural language. In: Proceedings of the IEEE Conference on Computer Vision and Pattern Recognition. pp. 4397--4406 (2018)

\bibitem{Ranftl2022}
Ranftl, R., Lasinger, K., Hafner, D., Schindler, K., Koltun, V.: Towards robust monocular depth estimation: Mixing datasets for zero-shot cross-dataset transfer. IEEE Transactions on Pattern Analysis and Machine Intelligence  \textbf{44}(3) (2022)

\bibitem{rombach2022high}
Rombach, R., Blattmann, A., Lorenz, D., Esser, P., Ommer, B.: High-resolution image synthesis with latent diffusion models. In: CVPR (2022)

\bibitem{rombach2021highresolution}
Rombach, R., Blattmann, A., Lorenz, D., Esser, P., Ommer, B.: High-resolution image synthesis with latent diffusion models (2021)

\bibitem{ronneberger2015u}
Ronneberger, O., Fischer, P., Brox, T.: U-net: Convolutional networks for biomedical image segmentation. In: International Conference on Medical image computing and computer-assisted intervention. pp. 234--241. Springer (2015)

\bibitem{saharia2022palette}
Saharia, C., Chan, W., Chang, H., Lee, C., Ho, J., Salimans, T., Fleet, D., Norouzi, M.: Palette: Image-to-image diffusion models. In: ACM SIGGRAPH 2022 Conference Proceedings. pp. 1--10 (2022)

\bibitem{saharia2022photorealistic}
Saharia, C., Chan, W., Saxena, S., Li, L., Whang, J., Denton, E.L., Ghasemipour, K., Gontijo~Lopes, R., Karagol~Ayan, B., Salimans, T., et~al.: Photorealistic text-to-image diffusion models with deep language understanding. NeurIPS  (2022)

\bibitem{sang2020single}
Sang, S., Chandraker, M.: Single-shot neural relighting and svbrdf estimation. In: ECCV (2020)

\bibitem{Shi2017CVPR}
Shi, J., Dong, Y., Su, H., Yu, S.X.: Learning non-lambertian object intrinsics across shapenet categories. In: Proceedings of the IEEE Conference on Computer Vision and Pattern Recognition (CVPR) (July 2017)

\bibitem{shi2017learning}
Shi, J., Dong, Y., Su, H., Yu, S.X.: Learning non-lambertian object intrinsics across shapenet categories. In: Proceedings of the IEEE conference on computer vision and pattern recognition. pp. 1685--1694 (2017)

\bibitem{shi2023gir}
Shi, Y., Wu, Y., Wu, C., Liu, X., Zhao, C., Feng, H., Liu, J., Zhang, L., Zhang, J., Zhou, B., Ding, E., Wang, J.: Gir: 3d gaussian inverse rendering for relightable scene factorization. Arxiv  (2023)

\bibitem{nerv2020}
Srinivasan, P.P., Deng, B., Zhang, X., Tancik, M., Mildenhall, B., Barron, J.T.: Nerv: Neural reflectance and visibility fields for relighting and view synthesis. CoRR  \textbf{abs/2012.03927} (2020), \url{https://arxiv.org/abs/2012.03927}

\bibitem{sun2023neuralpbir}
Sun, C., Cai, G., Li, Z., Yan, K., Zhang, C., Marshall, C., Huang, J.B., Zhao, S., Dong, Z.: Neural-pbir reconstruction of shape, material, and illumination (2023)

\bibitem{Tang_2023_ICCV}
Tang, J., Wang, T., Zhang, B., Zhang, T., Yi, R., Ma, L., Chen, D.: Make-it-3d: High-fidelity 3d creation from a single image with diffusion prior. In: Proceedings of the IEEE/CVF International Conference on Computer Vision (ICCV). pp. 22819--22829 (October 2023)

\bibitem{tang2023mvdiffusion}
Tang, S., Zhang, F., Chen, J., Wang, P., Furukawa, Y.: Mvdiffusion: Enabling holistic multi-view image generation with correspondence-aware diffusion. arXiv preprint arXiv:2307.01097  (2023)

\bibitem{NIPS2017Attention}
Vaswani, A., Shazeer, N., Parmar, N., Uszkoreit, J., Jones, L., Gomez, A.N., Kaiser, L.u., Polosukhin, I.: Attention is all you need. In: Guyon, I., Luxburg, U.V., Bengio, S., Wallach, H., Fergus, R., Vishwanathan, S., Garnett, R. (eds.) Advances in Neural Information Processing Systems. vol.~30. Curran Associates, Inc. (2017)

\bibitem{veach1995optimally}
Veach, E., Guibas, L.J.: Optimally combining sampling techniques for monte carlo rendering. In: Proceedings of the 22nd annual conference on Computer graphics and interactive techniques. pp. 419--428 (1995)

\bibitem{wang2023prolificdreamer}
Wang, Z., Lu, C., Wang, Y., Bao, F., Li, C., Su, H., Zhu, J.: Prolificdreamer: High-fidelity and diverse text-to-3d generation with variational score distillation. arXiv preprint arXiv:2305.16213  (2023)

\bibitem{wimbauer2022derendering}
Wimbauer, F., Wu, S., Rupprecht, C.: De-rendering 3d objects in the wild (2022)

\bibitem{wu2023nefii}
Wu, H., Hu, Z., Li, L., Zhang, Y., Fan, C., Yu, X.: Nefii: Inverse rendering for reflectance decomposition with near-field indirect illumination (2023)

\bibitem{fipt2023}
Wu, L., Zhu, R., Yaldiz, M.B., Zhu, Y., Cai, H., Matai, J., Porikli, F., Li, T.M., Chandraker, M., Ramamoorthi, R.: Factorized inverse path tracing for efficient and accurate material-lighting estimation. In: Proceedings of the IEEE/CVF International Conference on Computer Vision. pp. 3848--3858 (2023)

\bibitem{xie2023diffusion}
Xie, Y., Yuan, M., Dong, B., Li, Q.: Diffusion model for generative image denoising. arXiv preprint arXiv:2302.02398  (2023)

\bibitem{yao2022neilf}
Yao, Y., Zhang, J., Liu, J., Qu, Y., Fang, T., McKinnon, D., Tsin, Y., Quan, L.: Neilf: Neural incident light field for physically-based material estimation. In: European Conference on Computer Vision (ECCV) (2022)

\bibitem{yi2020leveraging}
Yi, R., Tan, P., Lin, S.: Leveraging multi-view image sets for unsupervised intrinsic image decomposition and highlight separation. In: Proceedings of the AAAI Conference on Artificial Intelligence. vol.~34, pp. 12685--12692 (2020)

\bibitem{yi2023weaklysupervised}
Yi, R., Zhu, C., Xu, K.: Weakly-supervised single-view image relighting (2023)

\bibitem{YuSelfRelight20}
Yu, Y., Meka, A., Elgharib, M., Seidel, H.P., Theobalt, C., Smith, W.: Self-supervised outdoor scene relighting. In: ECCV (2020)

\bibitem{yu19inverserendernet}
Yu, Y., Smith, W.A.: Inverserendernet: Learning single image inverse rendering. In: Proceedings of the IEEE/CVF Conference on Computer Vision and Pattern Recognition (CVPR) (2019)

\bibitem{yu2019inverserendernet}
Yu, Y., Smith, W.A.: Inverserendernet: Learning single image inverse rendering. In: Proceedings of the IEEE/CVF Conference on Computer Vision and Pattern Recognition. pp. 3155--3164 (2019)

\bibitem{10.1145/280814.280874}
Yu, Y., Malik, J.: Recovering photometric properties of architectural scenes from photographs. In: Proceedings of the 25th Annual Conference on Computer Graphics and Interactive Techniques. p. 207–217. SIGGRAPH '98, Association for Computing Machinery, New York, NY, USA (1998). \doi{10.1145/280814.280874}, \url{https://doi.org/10.1145/280814.280874}

\bibitem{Yu2022MonoSDF}
Yu, Z., Peng, S., Niemeyer, M., Sattler, T., Geiger, A.: Monosdf: Exploring monocular geometric cues for neural implicit surface reconstruction. Advances in Neural Information Processing Systems (NeurIPS)  (2022)

\bibitem{zhang2023neilfpp}
Zhang, J., Yao, Y., Li, S., Liu, J., Fang, T., McKinnon, D., Tsin, Y., Quan, L.: Neilf++: Inter-reflectable light fields for geometry and material estimation. In: ICCV (2023)

\bibitem{zhang2023neilf++}
Zhang, J., Yao, Y., Li, S., Liu, J., Fang, T., McKinnon, D., Tsin, Y., Quan, L.: Neilf++: Inter-reflectable light fields for geometry and material estimation. International Conference on Computer Vision (ICCV)  (2023)

\bibitem{zhang2022iron}
Zhang, K., Luan, F., Li, Z., Snavely, N.: Iron: Inverse rendering by optimizing neural sdfs and materials from photometric images. In: CVPR (2022)

\bibitem{physg2021}
Zhang, K., Luan, F., Wang, Q., Bala, K., Snavely, N.: {PhySG}: {I}nverse rendering with spherical gaussians for physics-based material editing and relighting. In: CVPR (2021)

\bibitem{zhang2018perceptual}
Zhang, R., Isola, P., Efros, A.A., Shechtman, E., Wang, O.: The unreasonable effectiveness of deep features as a perceptual metric. In: CVPR (2018)

\bibitem{Nefactor2021}
Zhang, X., Srinivasan, P.P., Deng, B., Debevec, P., Freeman, W.T., Barron, J.T.: Nerfactor. {ACM} Transactions on Graphics  (dec 2021). \doi{10.1145/3478513.3480496}

\bibitem{nerfactor}
Zhang, X., Srinivasan, P.P., Deng, B., Debevec, P., Freeman, W.T., Barron, J.T.: Nerfactor: Neural factorization of shape and reflectance under an unknown illumination. ACM Trans. Graph.  \textbf{40}(6) (dec 2021). \doi{10.1145/3478513.3480496}, \url{https://doi.org/10.1145/3478513.3480496}

\bibitem{zhang2022invrender}
Zhang, Y., Sun, J., He, X., Fu, H., Jia, R., Zhou, X.: Modeling indirect illumination for inverse rendering. In: CVPR (2022)

\bibitem{2022invrender}
Zhang, Y., Sun, J., He, X., Fu, H., Jia, R., Zhou, X.: Modeling indirect illumination for inverse rendering. In: CVPR (2022)

\bibitem{zhu2023i2}
Zhu, J., Huo, Y., Ye, Q., Luan, F., Li, J., Xi, D., Wang, L., Tang, R., Hua, W., Bao, H., et~al.: I2-sdf: Intrinsic indoor scene reconstruction and editing via raytracing in neural sdfs. In: Proceedings of the IEEE/CVF Conference on Computer Vision and Pattern Recognition. pp. 12489--12498 (2023)

\bibitem{zhu2022learning}
Zhu, J., Luan, F., Huo, Y., Lin, Z., Zhong, Z., Xi, D., Wang, R., Bao, H., Zheng, J., Tang, R.: Learning-based inverse rendering of complex indoor scenes with differentiable monte carlo raytracing. In: SIGGRAPH Asia 2022 Conference Papers. ACM (2022), \url{https://doi.org/10.1145/3550469.3555407}

\bibitem{zhu2023denoising}
Zhu, Y., Zhang, K., Liang, J., Cao, J., Wen, B., Timofte, R., Van~Gool, L.: Denoising diffusion models for plug-and-play image restoration. In: Proceedings of the IEEE/CVF Conference on Computer Vision and Pattern Recognition. pp. 1219--1229 (2023)

\end{thebibliography}

\end{document}